# Intention-Based Lane Changing and Lane Keeping Haptic Guidance Steering System

Zhanhong Yan, *Student member, IEEE*, Kaiming Yang, Zheng Wang, *member*, *IEEE*, Bo Yang, Tsutomu Kaizuka, Kimihiko Nakano, *member*, *IEEE*

*Abstract* — Haptic guidance in a shared steering assistance system has drawn significant attention in intelligent vehicle fields, owing to its mutual communication ability for vehicle control. By exerting continuous torque on the steering wheel, both the driver and support system can share lateral control of the vehicle. However, current haptic guidance steering systems demonstrate some deficiencies in assisting lane changing. This study explored a new steering interaction method, including the design and evaluation of an intention-based haptic shared steering system. Such an intention-based method can support both lane keeping and lane changing assistance, by detecting a driver's lane change intention. By using a deep learning-based method to model a driver's decision timing regarding lane crossing, an adaptive gain control method was proposed for realizing a steering control system. An intention consistency method was proposed to detect whether the driver and the system were acting towards the same target trajectories and to accurately capture the driver's intention. A driving simulator experiment was conducted to test the system performance. Participants were required to perform six trials with assistive methods and one trial without assistance. The results demonstrated that the supporting system decreased the lane departure risk in the lane keeping tasks and could support a fast and stable lane changing maneuver.

## I. Introduction

AUTOMATED driving is now expected to be a promising technology for improving traffic safety and efficiency [1]. In the Society of Automotive Engineers International J3016 document, an Automated Driving (AD) car is defined from automation level 0 (manual driving) to level 5 (fully automated driving) [2]. Level 1 to level 3 cars are usually considered as semi-automated cars, e.g., those equipped with a Driver Assistance System (DAS) for semi-automated driving. A considerable amount of effort has been expended to improve DASs to better serve drivers [3]. One aspect concerns the accurate prediction of driver intentions.

Predicting driver behavior can help a DAS to warn or interfere with the driver in advance, and thereby avoid potential accidents [4][5]. Many researchers have attempted to understand and predict driver behavior, especially with regard to Lane Changing (LC) maneuvers, as many accidents happen during this process [6][7]. Kumar et al. proposed an approach based on the Support Vector Machine that provided a multiclass probabilistic output as the prediction results [8]. Lethaus et al. proposed a method relying on gaze movement [9]. Dang's approach defined the time moment of lane crossing as labels, and their model could provide the exact time when the ego-vehicle would cross the lane boundary [10]. Yan et al. further explored Dang's method and proposed a deep learning-based model capable of predicting the end time of LC [11].

Great achievements have been made regarding driver behavior prediction, but few researchers have focused on combining driver intentions with DAS to achieve continuous steering assistance. Some intention-based devices, e.g., robot suits or exoskeletons, have proven effective in assisting users [12][13]. In [12], by measuring a user's electromyography signal, Lenzi explored an intention-based method that allowed a user and a powered system to share control authority for exoskeletons. The potential of a similar approach to DAS with a shared control steering vehicle can be expected to improve driving performance.

In shared control, a human driver and an automated system control an agent via a single operation input. It is expected to be an efficient method for automobile control, as realized by a Haptic Guidance Steering (HGS) system [14][15]. By exerting torque on an HGS wheel, the driver and the steering system can interact and share control authority. In relieving the driving workload, reducing lane departure risks, and other driving tasks, the HGS system demonstrates excellent performance relative to manual driving [16]. Many researchers have investigated methods for designing HGS systems and explored their effects via experiments. Mulder et al. assessed the performance of young and experienced drivers when driving with haptic steering in a curve negotiation task, and the results demonstrated that the safety boundaries were significantly

The research was supported by a Grant-in-Aid for Early-Career Scientists (No. 19K20318) from the Japan Society for the Promotion of Science, the China Scholarship Council (No.201706370201), and the Special Fund of Institute of Industrial Science, the University of Tokyo.

Zhanhong Yan, Zheng Wang, Bo Yang, Tsutomu Kaitsuka and Kimihiko Nakano are with the Institute of Industrial Science, the University of Tokyo, Tokyo, 112-0004, Japan (e-mail: {yanzhjob, z-wang, b-yang, tkaizuka, knakano}@iis.u-tokyo.ac.jp).

Kaiming Yang is with State Key Laboratory of Automotive Safety and Energy, Tsinghua University, Beijing, 100084, China. (e-mail: ykm739@126.com)

improved [17]. The influences of different degrees of haptic control on steering control were discussed in detail in [18]. Mars et al. collected driver performance and subject assessment data for six degrees of shared haptic steering control. Then, they proposed a method for assessing a suitable degree of shared control steering. The drivers provided with strong haptic assistance did not demonstrate improvements in steering behavior or subject feeling. The effectiveness of HGS in relieving the driver's mental and physical workload was proven in a contrast experiment with 12 participants [19]. In the experiment, a treatment session with an HGS system activated on a certain segment was compared with manual driving, to test the performance of the HGS in reducing passive fatigue. The HGS system was proven to be effective in improving driving performance. Indicators such as the mean absolute lateral error and time to lane crossing improved relative to the manual driving session. The HGS system also increased the physical demands according to the subjective feedback, and the increase caused a reduction in passive fatigue.

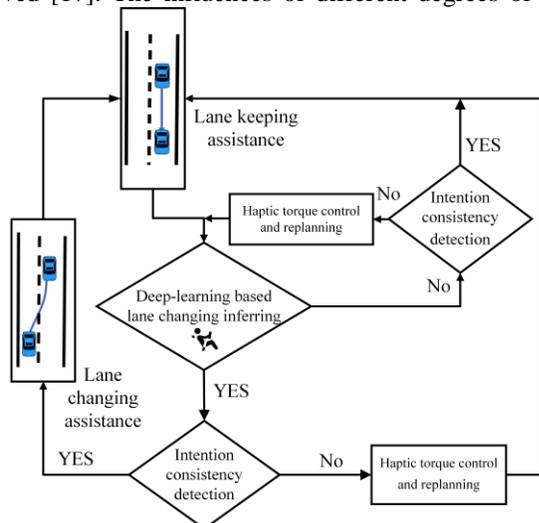

Fig. 1. Control flow chart of intention-based lane assistance haptic guidance steering (HGS) method

Previous research showed the potential of HGS application for vehicle control. However, compared with its application in a Lane Keeping (LK) task, the application of HGS systems for LC assistance is seldom discussed. The challenge lies in the fact that LK and LC have two opposite objectives. The realization of switching LK assistance to LC assistance makes the design of such a system difficult. A simple solution is leaving the decision to the human. As discussed in [20], a lane change decision supporting system would provide an environmental assessment to the driver, and he or she would decide the timing of the LC. However, such a system did not actually help the driver in steering behavior. In addition, researchers have explored HGS LC assistance using an additional trigger, but such design could be potentially annoying and inconvenient [21]. A more ideal approach is to automatically activate the assistive system. Tsoi proposed an HGS-based lane change assistance system, based on the time to lane crossing [22][23]. It fulfilled the requirements of both LK and LC, with acceptable feedback from drivers. However, this system did not consider the possibility of a false trigger in a LC, which could potentially lead to mistrust from drivers. In [24], Nishimura defined a cooperative status in an HGS and extracted a method of achieving LC. Based on the cooperative status, Nishimura's method reduced the haptic torque and left the LC task to the human driver. The drawback was that such a system did not actually provide a supporting torque for helping the driver changing lane.

In this research, we propose an Intention-Based Haptic Steering (IBHS) system for automatically switching from LK assistance to LC assistance, by detecting a driver's LC intention. The system is also capable of re-planning accordingly, by detecting whether the supporting system is acting the same as the driver's intention. The IBHS system is expected to improve driving performance as compared with manual driving, and this expectation is tested in a driving simulator experiment. Additionally, the experiment provides a basis for designing systems to better support drivers.

This paper is organized as follows. Section II illustrates the architecture of our IBHS system. Section III describes the driving simulator experiment including the scenario, conditions, apparatus, participants, procedure, and evaluation method. Section IV provides the statistical analysis results, and is followed by Section V, where the steering performance under different assistive methods is discussed. The conclusion is presented in Section VI.

## II. INTENTION-BASED HAPTIC GUIDANCE STEERING SYSTEM

We introduce an intention-based lane assistance haptic steering system for realizing smooth switch between LK and LC assistance. The purpose of this section is to introduce the architecture of the system.

### A. Overview

The architecture of our system is provided in Fig. 1. The system assumes that a driver is driving with an IBHS system which can adjust the assistive torque strength. Assuming that the HGS system is running in the LK assistive mode, at each time step, an LC intention inference model estimates the possibility of a driver's LC. At any time, the system can detect the cooperative state of the driver and haptic guidance system via an intent consistency module. The intent consistency module detects whether the guidance system has the same target trajectory as the driver. If an LC intention is detected and the system is acting consistently with the driver, the HGS system shifts into LC assistance mode and plans a changing trajectory to guide the driver. After changing, the system goes back to the LK mode. If an inconsistency is detected, the haptic torque gets adjusted, and a re-planning module may be activated to capture the driver's intention again.

### B. Lane changing intention and trajectory inferring

The initiation of the IBHS LC assistance relies on a lane changing prediction model. We assume the prediction model as $g(X_{t0})$. It outputs a result as:

For the current time step, whether the ego-vehicle will cross a lane boundary in the upcoming $m$ seconds.

According to the previous research [11], $m$ was set as three. The establishment of the model was based on a driving simulator experiment. In the experiment, nine participants (their ages ranged from 19 to 26, MEAN=22.8, SD=2.4) were invited to drive in an expressway scenario. The scenario included several LC events. A leading vehicle was set to be slower than the ego-vehicle to make the drivers changing lane. In the experiment, 432 times of LC (236 times from left to right, 196 times from right to left) were observed. The data were sampled at a rate of 60 Hz. A detailed description of the recorded data can be found in Table I.

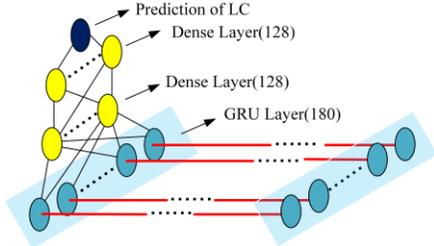

Fig.2. Unrolled GRU layer with 180 units, the top layer gives the prediction of driver's lane changing intention

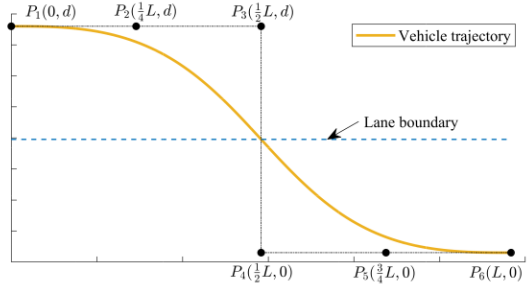

Fig. 3. 5-th Bézier curve lane changing (LC) trajectory, where $d$ is road width. $L=V_x \cdot \Delta T_{LC}$ decides the speed of the LC

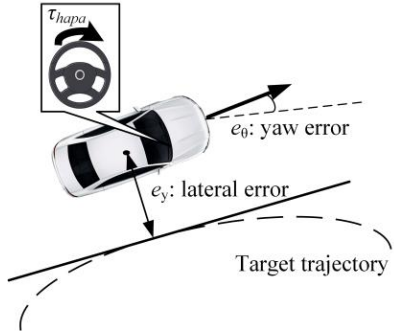

Fig. 4. Diagram of single preview-point method for haptic guidance torque

TABLE I
LANE CHANGING INTENTION DETECTION INPUT

| Symbol | Description | unit |
|---|---|---|
| Head | left/right-rotation of the head, the head yaw angle | [-1,1], normalized |
| $a$ | Longitudinal acceleration | m/s² |
| $v$ | Longitudinal velocity | m/s |
| $\theta_{sw}$ | Steering wheel angle | rad |
| $d_{adj}$ | Lateral distance to the adjacent lane | m |
| $\psi$ | Vehicle yaw angle | rad |

TABLE II
PARAMETERS OF STEERING CONTROLLER DESIGN

| Symbol | Description | unit |
|---|---|---|
| $J_{eq}$ | Steering moment of inertia about steering column | kg·m² |
| $B_{eq}$ | Steering moment of damping about steering column | Nm·s/rad |
| $K_{Fz}$ | Steering resistance coefficient | Nm/rad |
| $\alpha$ | Side slip angle of front tire | rad |

Assuming the current time step is $t_0$, the data in the past $k$ time steps as a vector $X_{t_0}$ is:

$$X_{t_0}=[Head_{t_0-k+1}, a_{t_0-k+1}, v_{t_0-k+1}, \theta_{sw\,t_0-k+1}, d_{adj\,t_0-k+1}, \psi_{t_0-k+1},$$
$$Head_{t_0-k+2}, \ldots, d_{adj\,t_0}, \psi_{t_0}]^T \quad (1)$$

Where $k = 180$. Because the sampling rate was 60 Hz, $k = 180$ represented that $X_{t_0}$ covered the data in the past three seconds. The prediction model can be then described as:

$$g(X_{t_0}) = \begin{cases} 1, \text{ the ego-vehicle will cross the boundary} \\ 0, \text{ the ego-vehicle will remain in the lane} \end{cases} \quad (2)$$

Since drivers tend to perform a series of behaviors, e.g., observing side mirrors or accelerating before changing a lane, the short period before a LC maneuver can cover significant information for the intention prediction. The Gated Recurrent Unit (GRU) is an ideal tool for the prediction task, due to its ability to discover dependency in time series[25].

Based on the experimental data, 19000 samples were used for training and 3800 samples were used for validation. Three participants were selected and their last 30 mins driving data, respectively (about 2200 samples) were set as the test data. We established a GRU based network as shown in Fig. 2. The test results suggested that at any time, our model could predict whether a driver wants to change a lane at an accuracy of 94.8%. Additional details of the model can be found in [11].

Once the changing assistance is launched, the trajectory planning model then plans a changing trajectory based on a 5-th Bézier curve referred from [26] and [27] as the inferred LC trajectory, as shown in Fig. 3. Compared with search-based or probabilistic (e.g., random tree) methods that are inefficient in time-critical situations, the Bézier curve is geometrically based, and requires less computation than a clothoid trajectory [28]. The length of the entire LC process $L$ is defined according to the current speed when the LC intention is detected. $\Delta T_{LC}$ determines the speed of one LC. A shorter $\Delta T_{LC}$ causes a sharper turn, but shortens the time to change:

$$L = V_x \cdot \Delta T_{LC} \quad (3)$$

### C. Vehicle lateral control

The haptic guidance assistance torque applied on the steering wheel is determined as follows:

$$\tau_{hapi} = J_{eq}\ddot{\theta}_{sw} + B_{eq}\dot{\theta}_{sw} + K_{Fz}\theta_{sw} - \tau_{dr}' + \tau_{dis} \quad (4)$$
$$\tau_{dr}' = K_y e_y + K_{yd}\dot{e}_\theta + K_\theta \theta_{sw} + K_\alpha \alpha \quad (5)$$
$$\tau_{hapa} = K_h \tau_{hapi} \quad (6)$$

In the above, $\tau_{hapi}$ is the haptic guidance torque instruction value. It is amplified by the gain $K_h$ (ranging from 0 to 1) and is converted to the actual haptic torque $\tau_{hapa}$. Tuning gain $K_h$ represents the spectrum from fully automated driving to manual driving when it changes from "1" to "0". The tuning gain $K_h$ enables the IBHS system to tune its output adaptively, according to the state of intention consistency. $\tau_{dr}'$ denotes the estimated torque from the human, according to a single-point preview method (see Fig. 4). $\tau_{dis}$ represents the equivalent

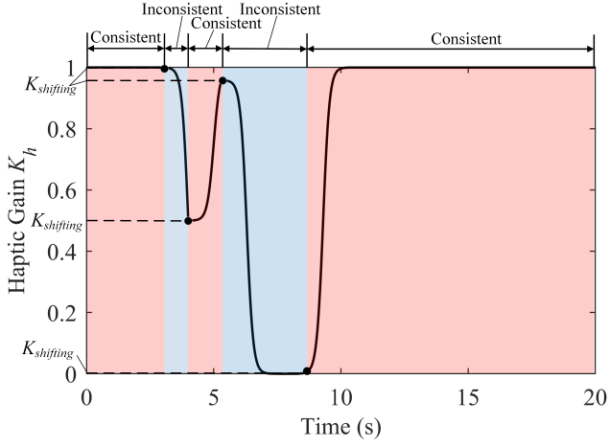

Fig. 5. Example of gain $K_h$ adjustment

TABLE III
INTENTION CONSISTENCY DETECTING MECHANISM

| $S_c<0$ | | $S_c>0$ |
|---|---|---|
| $\|W_{hapi}-W_{dr}\|=\beta>\delta$ | $\|W_{hapi}-W_{dr}\|=\beta<\delta$ | |
| Inconsistent | Consistent | Consistent |

$\delta$: Threshold to whether inconsistency happens or not.

friction torque in the steering system and the self-aligning torque from the tire force. $K_y$, $K_{yd}$, $K_\theta$, and $K_\alpha$ are the corresponding gains of $e_y$, $e_\theta$, $\theta_{sw}$, and $\alpha$. The remaining parameters are described in Table I and Table II. Additional details regarding the controller design and the method for estimating $\tau_{dr}$ can be found in [29] and [30].

*D. Intention consistency detection*

The intention consistency detection model defines a method for judging whether the IBHS system and driver are driving towards the same target trajectory. In [31], Saito et al. proposed a method that depended on comparing efforts on the steering wheel from the haptic guidance system and driver to determine cooperative states, known as "pseudo-work". The main drawback of the pseudo-work approach is that it must rely on the vehicle lateral motion response $v_y$. If a driver intends to stay in LK mode but the HGS mistakenly activates to support a LC maneuver, the ego-vehicle may only have small vehicle lateral motion. Saito's method cannot distinguish such circumstances. However, inspired by this, we further characterize the intention consistency features and propose a modified pseudo-work method, as shown in (5) to (7) as follows:

$$W_{hapi} = \frac{1}{\Delta T}\int_{t-\Delta T}^{t} \tau_{hapi}(t)\dot{e}_\theta(t)dt \quad (7)$$

$$W_{dr} = \frac{1}{\Delta T}\int_{t-\Delta T}^{t} \tau_{dr}(t)\dot{e}_\theta(t)dt \quad (8)$$

$$S_c = \tau_{hapi} \cdot \tau_{dr} \quad (9)$$

Here, $\Delta T$=1 s. The nature of the inconsistency is owing to the disparate target trajectories of the driver and the HGS system. Therefore, the modified method relies on $\dot{e}_\theta$ (see Fig. 4) to measure the degree of departure between the two target trajectories. $S_c$ is for narrowing down the detection zone. The calculation of $W_{hapi}$ and $W_{dr}$ does not require actual vehicle lateral movement and can fit more circumstances. Based on Table III, the intention consistency method is defined and applied.

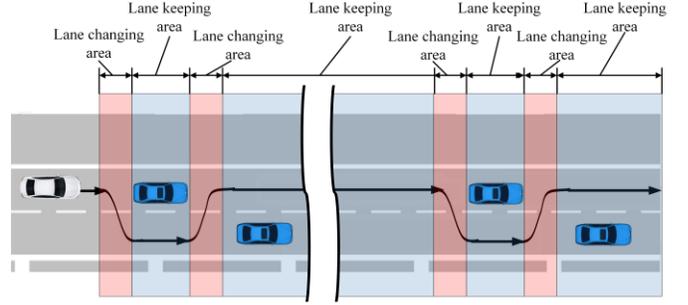

Fig. 6. Lane changing area and lane keeping area in the experiment scenario

*E. Haptic torque strength control and trajectory re-planning*

When an inconsistency is detected, the gain $K_h$ gets reduced, to allow control of the actual haptic guidance torque $\tau_{hapa}$. This adjustment temporarily allows the driver to have more steering control authority and leaves the IBHS system to re-plan and capture the driver's intention again. A strength control method is used to realize a smooth control authority transfer, as shown below:

When the intentions are in an inconsistent state, $K_h$ tends to drop to 0, as follows:

$$K_h = -\frac{K_{shifting}}{2}\tanh[\lambda \cdot (t-\gamma)]+\frac{K_{shifting}}{2} \quad (10)$$

When the intentions are in a consistent state, $K_h$ tends to recover to 1, as follows:

$$K_h = \frac{1-K_{shifting}}{2}\tanh[\lambda \cdot (t-\gamma)]+\frac{1+K_{shifting}}{2} \quad (11)$$

In the above, $\lambda = 4$, $\gamma = 1$, and $t \in [0,+\infty)$ is the time duration for remaining in the current state. The hyperbolic tangent functions are monotonic in the real domain and their ranges are limited in (0,1). Every time the state switches from consistent to inconsistent or vice versa, $t$ restarts the time. $K_{shifting} \in (0,1)$ takes the value of $K_h$ at the consistency state switching moment.

Equations (10) and (11) form a continuous curve for haptic strength adjustment. The parameters $-\frac{K_{shifting}}{2}$, $\lambda$, $\gamma$, $\frac{K_{shifting}}{2}$ in Equations (10) and $\frac{1-K_{shifting}}{2}$, $\lambda$, $\gamma$, $\frac{1+K_{shifting}}{2}$ in Equations (11) are to guarantee that the haptic guidance torque remains in (0,1) .When state switching, and $t = 0$, the starting points of (8) and (9) are as follows:

$$K_h= -\frac{K_{shifting}}{2}\tanh(-4)+\frac{K_{shifting}}{2} \approx K_{shifting} \quad (12)$$

$$K_h= \frac{1-K_{shifting}}{2}\tanh(-4)+\frac{1+K_{shifting}}{2} \approx K_{shifting} \quad (13)$$

The above ensure no abrupt change on the torque strength, and $K_h$ will finally return to 1 or drop to 0 over time. Fig. 5 presents a $K_h$ adjustment example in the context of a system state switching. This method is intended to adaptively control the strength when an inconsistency occurs, and to reduce steering shaking when changes in torque strength.

Trajectory re-planning occurs when $K_h$ reaches 0. The HGS system always re-plans a guidance trajectory to the current lane and assumes that the ego-vehicle is in the LK assistive mode. Owing to the gain control, the haptic guidance torque will be small when an inconsistency occurs. This allows the driver to completely control the vehicle for a short time. The re-planning strategy is based on the concept that the driver may have better insight into the best solutions to handle the current situation compared with the assistive system, provided the control authority can be safely transferred.

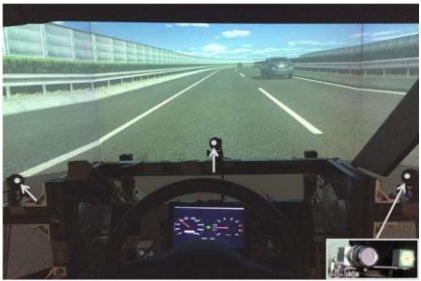

Fig. 7. Fixed driving simulator and Smart Eye Pro eye tracking system. Reproduced from [32]

TABLE IV
EXPERIMENT DESIGN

| Condition | Factor one Haptic torque strength (When in consistent) | Factor two lane changing (LC) assistance |
|---|---|---|
| 1 Manual | No | No |
| 2 Strong-Rapid | Strong, $\tau_{hapa} = K_h \tau_{hapi}$ | Rapid, $\Delta T_{LC} = 4$ s |
| 3 Strong-Normal | Strong, $\tau_{hapa} = K_h \tau_{hapi}$ | Normal, $\Delta T_{LC} = 6$ s |
| 4 Strong-Gentle | Strong, $\tau_{hapa} = K_h \tau_{hapi}$ | Gentle, $\Delta T_{LC} = 8$ s |
| 5 Weak-Rapid | Weak, $\tau_{hapa} = 0.4 \cdot K_h \tau_{hapi}$ | Rapid, $\Delta T_{LC} = 4$ s |
| 6 Weak-Normal | Weak, $\tau_{hapa} = 0.4 \cdot K_h \tau_{hapi}$ | Normal, $\Delta T_{LC} = 6$ s |
| 7 Weak-Gentle | Weak, $\tau_{hapa} = 0.4 \cdot K_h \tau_{hapi}$ | Gentle, $\Delta T_{LC} = 8$ s |

## III. EXPERIMENT

### A. Scenario

The experimental scenario was a straight two-lane expressway (lane width: 3.5 m, total length: 8 km). The drivers would meet four LC events, including one LC to the right, and one free change to the left (see Fig. 6). All leading vehicles were set to be randomly slower (5 km/h to 1 5 km/h) than the ego-vehicle to cause the driver to change lanes. For the rest of the course, the drivers were requested to remain in the lane.

### B. Conditions

All participants drove seven trials to investigate two factors, as shown in Table IV. In manual driving conditions, the subjects drove without assistance. In conditions 2 to 7, the subjects drove with the IBHS system and different assistance methods. All participants were required to hold the steering wheel during each entire trial.

Factor one was intended to explore how the assistive torque strength affected driving behavior. A fine-tuned level of shared control is beneficial to achieving high-quality cooperation between a driver and the steering system [18][32]. In [18], the researchers explored the performance of drivers with strong assistive torque (automated lane tracking), weak assistive torque (steering direction guidance), and no assistive torque (manual driving). The results demonstrated that high-level automation in shared control could be annoying or intrusive to the drivers. Therefore, the effects of the torque strengths in IBHS system would be investigated in our experiment. The data collected from methods 2, 3, and 4 were treated as the "Strong" group, and that from methods 5, 6, and 7 were treated as the "Weak" group. For Strong conditions, $K_h$ was set to be 1 (when the system and the driver moved consistently), representing that the assistive torque could support fully automated steering. For Weak conditions, $K_h$ was set to not exceed 0.4. Such a restriction would only support a guidance steering torque, and drivers were still required to exert torque on the steering wheel to drive.

Factor two was intended to investigate how to better support the driver to change a lane. According to [33], a LC maneuver takes approximately six seconds. In [22], the researchers proposed a haptic steering lane changing assistance system. From fast to slow, they tested the three LC trajectories. Based on the current speed, the guidance LC trajectories would be followed to guide the drivers to the adjacent lane in 4, 6, 8 seconds. Therefore, factor two was set as 4, 6, 8 seconds to find out a suitable LC speed. The driver acceptance of a fast (4 s, faster than natural manual driving) or a slow (8 s, slower than natural manual driving) LC methods would be investigated. The data collected from methods 2 and 5 were grouped into a "Rapid" group. Methods 3 and 6 were the "Normal" group and methods 4 and 7 were the "Gentle" group. The change of $\Delta T_{LC}$ causes the target LC trajectory to be rapid or slow.

For the driver, there were no restrictions on steering and accelerating, but they were suggested to drive at a speed of approximately 70 km/h.

### C. Apparatus

The experiment was conducted in a stationary driving simulator, as shown in Fig. 7. A 140° field-of-view of driving scene was generated by three projectors. The self-aligning, friction, and assisting steering torque were emulated via an electric steering system, and the raw data in the experiment were recorded at a sampling rate of 60 Hz. The maximum assisting steering torque was restricted so that the driver can overrule it if necessary.

A gaze-tracking system (Smart Eye AB, Sweden) was employed to measure the driver's head and eye movements. The "Smart Eye" system consisted of three cameras placed in front of the subject and caused no physical burden to the participants.

### D. Participants

A dozen healthy volunteers (two females and ten males) whose ages ranged from 20 to 35 (mean = 24.3 and SD = 3.5) were recruited to participate in the experiment. All participants had normal or corrected-to-normal vision and a valid Japanese driver's license (mean driving experience of 4.4 years and SD = 3.9). They also received monetary compensation for their participation and were informed that they were free to withdraw from the study at any time for any reason. The study received ethical approval by the Office for Life Science Research Ethic and Safety, the University of Tokyo.

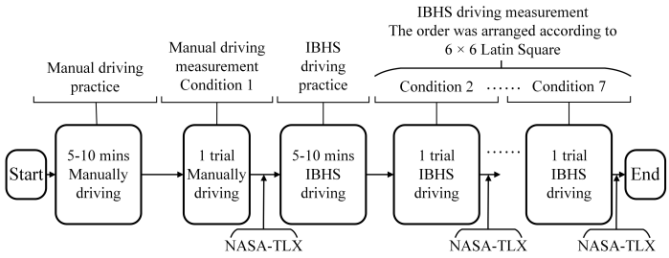

Fig. 8. Experimental procedure

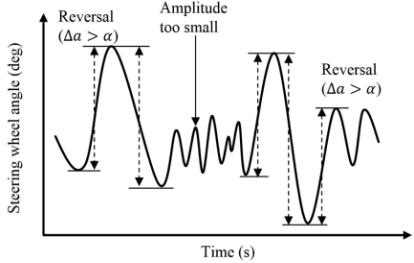

Fig. 9. Steering wheel reversal rate (SWRR) calculation, adapted from [35], reproduced from [37].

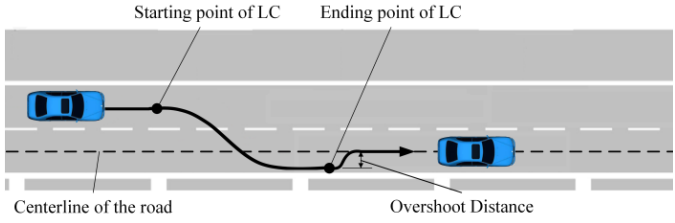

Fig. 10. LC start and end, and overshoot distance measurement

### E. Procedure

The procedure for the experiments is shown in Fig. 8. Before the procedure started, all participants were ensured to be naive to the purposes of the experiment. Each participant signed a consent form explaining the procedure of the experiment and filled in a questionnaire regarding their basic information at the beginning of the experiment. The experiment started with manual driving, consisting of one practice trial lasting approximately 5–10 min (more practices were possible if proposed), and an experimental manual driving trial. The participants were asked to follow Japanese traffic rules during the entire experiment. Then, each subject performed IBHS driving trials, which consisted of a practice trial and six experimental trials (each corresponding to one lane assistance method). The order of the IBHS driving trials was arranged according to the Latin Square [34]. Following each trial, a NASA Task Load Index (NASA-TLX) was calculated to assess the participants' subjective workloads, in which a higher score represented a heavier workload [35]. The NASA-Task load index consists of six items: Mental Demand, Physical Demand, Temporal Demand, Performance, Effort and Frustration.

### F. Evaluation

The evaluation focused on assessing the driving performance under LK assistance and LC assistance (see Fig. 6 for the measurement areas of LK and LC). The driver's overall mental workload was determined through the NASA-TLX, and his/her physical load was calculated as the Root Mean Square (RMS) of the driver's applied torque. The intention catching performance was also discussed. We determined the end and the beginning of a LC maneuver by detecting the yaw angle of the ego-vehicle.

For LK assisting performance, we measured the Standard Deviation of the Lane Position (SDLP) and the Steering Wheel Reversal Rate (SWRR). The SDLP is calculated as follows:

$$\text{SDLP} = \sqrt{\frac{1}{N-1}\sum_{i=1}^{N}(x_i - \mu)^2} \qquad (12)$$

Here, $x_i$ is the lateral position of the ego-vehicle, $N$ is the number of collected samples taken in each driving course, and $\mu$ is the mean lateral position of the vehicle. The SWRR is defined as the number of changes in the steering wheel direction per minute [36]. Fig. 9 illustrates the counting of steering reversal behavior; $\alpha=3°$ refers to the threshold for measuring larger steering maneuvers.

For LC assisting performance, we measured the LC duration time and overshoot distance (for undershoot distance, we considered the absolute value), as shown in Fig. 10. Additionally, the RMSs of the steering wheel angular velocity and peak steering wheel angle during the changing process were calculated.

The data analysis was conducted via a repeated one-way Analysis of Variance (ANOVA) to investigate the statistical effects of the IBHS systems. Post-hoc $t$-tests, corrected using a Benjamini-Hochberg procedure for the method factors, were used to determine the significant differences [37]. The significance level was set as 0.05.

## IV. RESULTS

### A. Overview of measured signals

Fig. 11 illustrates experimental signals under conditions of manual driving and IBHS-assisted driving (Strong-Normal), respectively. A greater steering velocity was observed as compared with manual driving during a LC maneuver. The average trajectories in LC maneuvers are presented in Fig. 12.

### B. Lane keeping

#### 1) Standard deviation of lane position

Fig. 13 illustrates the results for the SDLP for all driving conditions, with error bars representing 0.95 confidence intervals. As the second factor, $\Delta T_{LC}$, does not affect the system performance in LK performance, the analysis focuses on comparing how the strength affected the driving performance. The ANOVA showed that the SDLP was significantly different among the groups ($F$ (2,81) = 9.61, $p < 0.001$). The $t$-tests revealed that the SDLPs for Strong and Weak assisting torque were significantly lower than that for the Manual condition ($p < 0.001$, $p = 0.003$, respectively). However, no significance was found between the different strengths ($p = 0.087$).

#### 2) Steering wheel reversal rate

The results for the SWRR along the LK areas is shown in Fig. 14. The SWRR was not significantly different in the different driving conditions ($F$ (2,81) = 1.24, $p = 0.293$). The

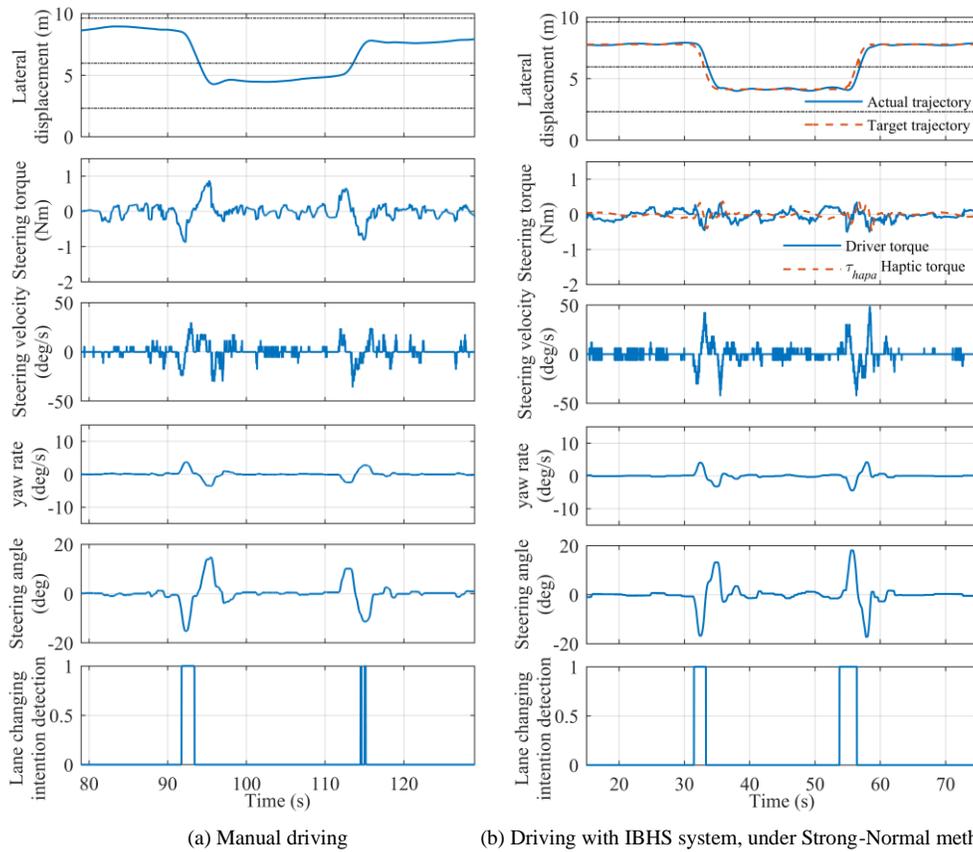

(a) Manual driving  (b) Driving with IBHS system, under Strong-Normal method
Fig. 11. Overview of measured signals collected from subject No. 6. Dot-dashed lines in "Lateral displacement" represent lane boundaries. The "Lane changing intention detection" equals 1 when the system detects that the driver is about to change lanes.

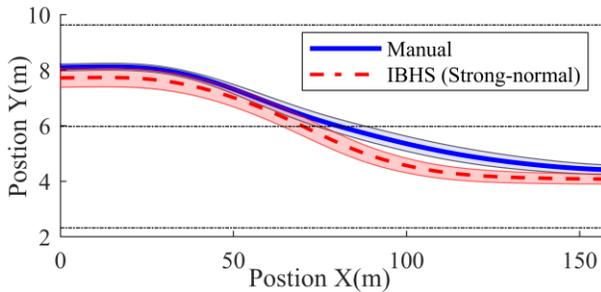

Fig. 12. Average vehicle trajectory including Manual and Strong-Normal with 0.95 confidence intervals of all participants. Dot-dash lines represent lane boundaries

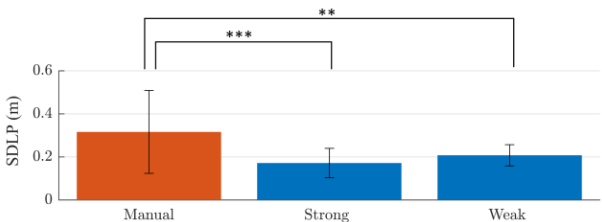

Fig. 13. Standard deviation of lane position (SDLP). Error bars denote 0.95 confidence intervals. Where ***: $p < 0.001$, **: $p < 0.01$, *: $p < 0.05$. (the same as below).

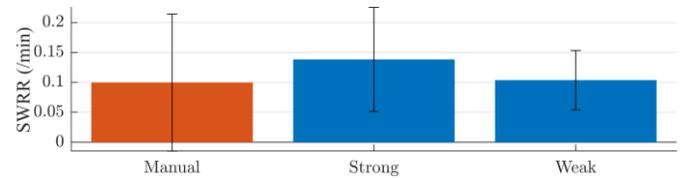

Fig. 14. Steering wheel reversal rate (SWRR).

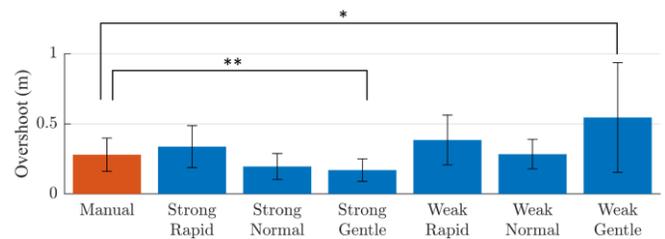

Fig. 15. Overshoot distance when finishing lane changing

insignificance suggested that the demand for steering direction correction was satisfactory under conditions of manual driving.

### C. Lane changing

#### 1) Overshoot

As shown in Fig. 15, overshoots were observed differently regarding statistical significance in the Manual condition and IBHS assistance ($F$ (6,652) = 7.23, $p < 0.001$) conditions. However, the $t$-tests between the Manual and different IBHS systems indicated that only the Strong-Gentle method significantly reduced the overshoot distance ($p = 0.002$). The Strong-Normal method almost showed statistical significance in decreasing the overshoot distance. Its $t$-test result ($p = 0.051$) was close but not strong enough. In contrast, the Weak-Gentle method demonstrated significance in increasing the overshoot distance ($p = 0.031$).

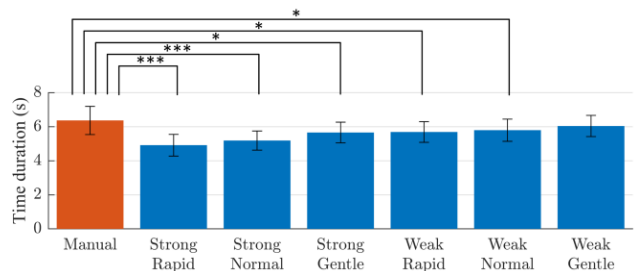
Fig. 16. Lane changing duration time

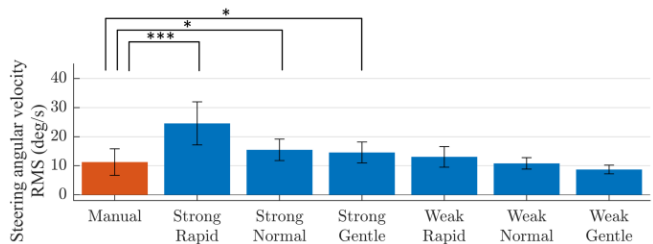
Fig. 17. Steering wheel angular velocity when shifting

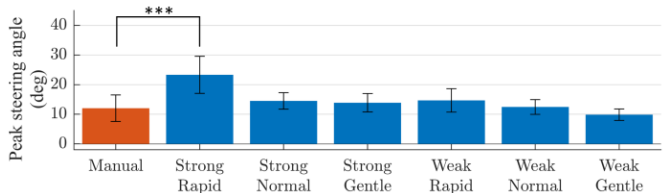
Fig. 18. Peak steering wheel angle when shifting

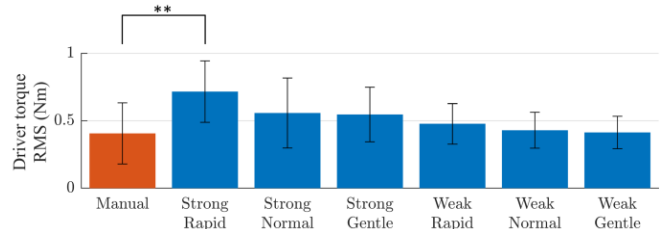
Fig. 19. Driver physical workload, estimated via RMS of driver torque

A two-way ANOVA test for the different IBHS methods illustrated the significance of the torque strengths ($F(1,559) = 17.97$, $p < 0.001$), and $\Delta T_{LC}$ indicated the speed of the LC ($F(2,559) = 3.96$, $p = 0.020$). Significance was also found in the interaction ($F(2,559 = 6.6, p = 0.002$). It can be observed that a strong assisting torque was more effective than a weak one in reducing the overshoot distance. There was also a tendency that when supporting a strong torque, the overshoot was reduced when $\Delta T_{LC}$ became longer.

*2) Duration*

The mean time duration of a LC is plotted in Fig. 16. It can be observed that the duration showed a significant reduction after supporting steering torque ($F(6,652) = 9.03$, $p < 0.001$). The *t*-tests between the Manual and different IBHS systems indicated that except for the Weak-Gentle method ($p = 0.208$), all methods showed significance in shortening the LC duration.

The two-way ANOVA test for IBHS methods revealed the significance in changing the LC duration among torque strengths ($F(1,559) = 21.63$, $p < 0.001$), and $\Delta T_{LC}$ revealed the speed of LC ($F(2,559) = 6.4$, $p = 0.002$). However, no significance was found in the interaction ($F(2,560) = 0.81$, $p = 0.447$). It can be seen there was a tendency for the duration to be reduced when $\Delta T_{LC}$ became shorter. Thus, a strong assisting torque is more effective than a weak one in shortening the time duration.

*3) Steering angular velocity*

The mean steering wheel angular velocity during LC is presented in Fig. 17. The RMS of the steering angular velocity was significantly different under conditions of manual and haptic guidance assisting driving ($F(6,652) = 24.39$, $p < 0.001$). However, only the Strong group showed significance in increasing the RMSs of the steering angular velocity (Rapid: $p < 0.001$, Normal: $p = 0.014$, Gentle: $p = 0.049$).

For the different IBHS methods, the two-way ANOVA revealed significance in changing the steering velocity among assisting torque ($F(1,559) = 75.86$, $p<0.001$) and $\Delta T_{LC}$ ($F(2,559) = 26.84$, $p < 0.001$). Significance also was found for the interaction ($F(2,559) = 0.81$, $p = 0.002$). This conclusion agrees with the duration changing tendency, as fast steering behaviors generally shorten the LC duration.

*4) Peak steering wheel angle*

Fig. 18 shows the peak steering wheel angle time when changing lanes. For the ANOVA test in all conditions, $F(6,652) = 19.71$, $p < 0.001$. However, the *t*-tests between the Manual condition and the different IBHS systems suggest that only the Strong-Rapid method ($p<0.001$) increased peak angle. There was a similar decreasing tendency in the peak steering angle as compared with the duration.

For the different IBHS methods, there was a significant effect from changing the peak angle among the torque strengths ($F(1,559) =41.09$, $p<0.001$), and $\Delta T_{LC}$ ($F(2,559) = 31.79$, $p < 0.001$). The interaction was also found with significance ($F(2,559) = 6.5$, $p = 0.002$).

### D. Driver workload

*1) Physical workload*

Fig. 19 shows the average RMSs of the driver torque when driving along an entire course. There was a significant effect from the physical workload ($F(6,77) = 6.25$, $p < 0.001$). The *t*-tests between the Manual and different IBHS systems suggested that only the Strong-Rapid method increased the required torque for lateral control ($p<0.002$).

For the different IBHS methods, there was a significant effect from changing driver torque among the torque strengths according to a two-way ANOVA ($F(1,66) = 26.02$, $p < 0.001$). However, for $\Delta T_{LC}$ ($F(2,66) = 2.57$, $p = 0.084$) and the interaction ($F(2,66) = 0.15$, $p = 0.859$), no statistically significant effects were found regarding burdening the driver's physical workload.

*2) Mental workload*

As shown in Fig. 20, the average overall mental workload was estimated via the NASA-TLX score. No statistical difference was found ($F(6,77) = 1.66$, $p = 0.1421$). Fig. 21 presents the NASA-TLX scores of the Manual and the Strong-Normal methods. Though no statistical difference was found, there was a tendency that the Strong-Normal method reduced workload in all indexes except the Frustration.

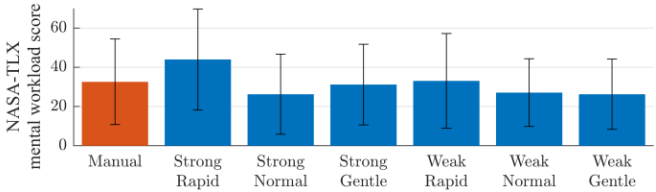

Fig. 20. Driver mental workload, estimated via NASA task load index (NASA-TLX) score

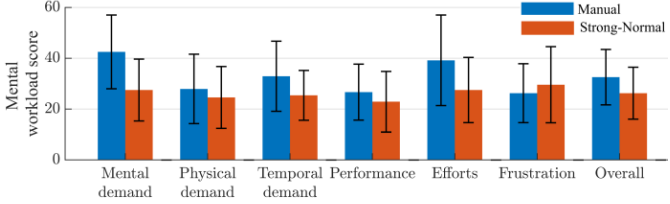

Fig. 21. NASA-TLX scores of Manual and Strong-Normal

### E. Intention consistency detecting and re-planning

Fig. 22. presents an example of how the system handles inconsistent intentions. In the example, the intention prediction system determined a wrong LC maneuver, and tried to guide the driver to an adjacent lane. With the target trajectory deviating from the current lane, the modified pseudo-work from both agents grew, and reached the threshold. The system then went into an inconsistent state. The haptic guidance gain $K_h$ therefore decreased according to Equation (8), leaving the driver to temporarily control the vehicle. When $K_h$ dropped close to 0, trajectory re-planning was initiated. The system went back to LK assistance, and $K_h$ recovered to 1 without further negative impacts.

During the entire experiment, each time an inconsistency was detected, the system worked properly, and adjusted $K_h$ correctly to follow the driver.

In summary, intention inconsistency happened more frequently in the Rapid group than in the Normal and Gentle groups, especially when a driver just finished changing. It was likely that excessively rapid changing caused mismatches with the driver's will, and he/she tried to overrule the steering wheel, resulting in the $K_h$ adjustment.

## V. DISCUSSION

### A. Assisting performance

For LK assistance, the conducted experiment explored how the torque strength affected driving behavior. From the results, the SDLP was significantly reduced when driving with an IBHS system as compared with manual steering. A decrease in the SDLP generally indicates a more stable LK behavior, with less lane departure risk. However, the post-hoc *t*-test indicated that there was no difference in the SDLP among the different strength conditions. In addition, the SWRR showed no difference in comparison to manual driving. The SWRR implies a higher departure risk when the driver shows a low LK performance [36]. The steering performance was satisfying under the condition of manual driving.

To sum up, these findings agree with previous research in related areas [18][19][38], suggesting that the addition of an intention detection and intention inconsistency module has no side effect on the LK performance.

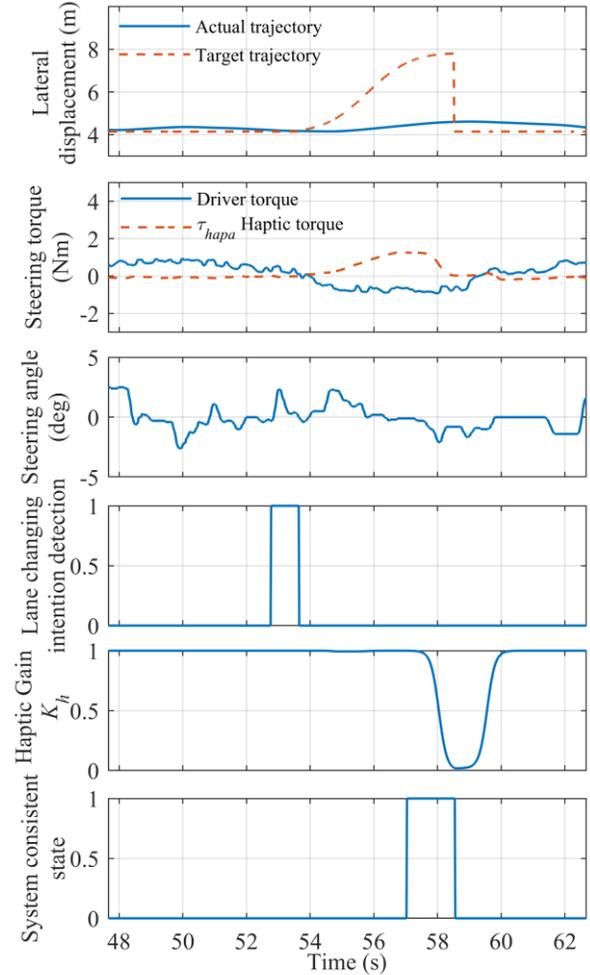

Fig. 22. Overview of intention inconsistency handling collected from subject No. 7 with Strong-Slow method assistance. The "System consistent state" equals one when the system and the driver move inconsistently

Regarding LC assistance, there was a tendency in the Strong group for the overshoot distance to reduce when driving with a slower changing trajectory. The Strong-Gentle method showed significance in reducing the overshoot distance. The Strong-Normal group almost showed a similar tendency. However, for the Weak group, the overshoot distance was increased when the changing trajectory became slower. It is likely that the weak guidance torque could not perform well in reducing the overshoot distance. A small overshoot generally means that the driver can easily get back into LK with little steering wheel readjustment. In shortening the time duration of LC, the IBHS also proved to be statistically effective, except when using the Weak-Gentle method.

For steering control performance, an angular velocity reduction tendency similar to that in overshoot can be observed in both the Strong and Weak groups. Such a tendency is in accordance with the shortening of the duration, as a faster steering velocity generally requires less time to finish a LC. However, the results showed that only the Strong-Rapid method significantly increased the peak angle. The peak angle can indicate the stability of a steering operation, as discussed in [31][32]. In a normal situation, a larger peak value can sometimes indicate a sharp turn and intense vehicle movements, which can potentially cause uncomfortable sensations. In this sense, the two Rapid methods may prove to be inappropriate.

To sum up, the Strong-Normal method proved to be the best combination of all methods, considering its effects in reducing the overshoot distance and time duration. It will allow a fast and guided LC, without causing instability in the steering operation.

### B. Driver workload

A similar reduction tendency can be observed in the physical workload in both the Strong and Weak groups. Only the Strong-Rapid method significantly increased the physical workload relative to the Manual method. The reason for the increased load may be presumed as the extra efforts required when the driver and vehicle targets are slightly different. This happens frequently when the system is intended to achieve fast LC assistance. Rapid methods require a fast turning, which might be unexpected for drivers. Therefore, the driver had to use more effort to overrule the wheel and slow the steering maneuver. In contrast, the Normal and Gentle methods did not show such effects. A slow LC trajectory may be fit to drivers' habits, in agreement with the research in [33]. Meanwhile, as compared to the Strong methods, the Weak methods required less effort to overrule; therefore, they caused no significant increase in workload.

In addition, the Strong-Normal method showed a tendency to reduce workload in all the indexes except the Frustration. The tendency suggested that the drivers could steer easier and more comfortable than manual driving. As for the feeling of frustration, a potential explanation could be that when the drivers disagreed with the guidance torque, it might cause the drivers to resist and cause frustration [38].

### C. Overall

The above results indicate the effectiveness of proposed system in assisting drivers. Among all the methods, the Strong-Normal method demonstrated the best performance in both LK and LC assistance tasks. It can reduce the lane departure risk and support fast and stable LC. Although such a method may increase the physical workload, Wang's research in [19] also indicated that increased physical control demands may help to reduce passive fatigue. The existence of the intention catching module ensures that the system can always follow the driver's desired target trajectory.

However, some research also showed that a strong guidance torque may not be the best way to support steering control, which is the opposite of our experiment results [18][32]. This difference may be owing to the intention-based steering system, as our experimental conditions were not the same as in [18] and [32]. It can be assumed that if the haptic system can act accurately regarding the driver's desired trajectory, even a strong torque may not cause discomfort.

Many researchers have explored the haptic guidance control for vehicle lateral control [18][19]. HGS has been proven to be effective in reducing lane departure risks in LK tasks. However, for a LC assistive task, the switch between two modes can become a problem, as LC has an opposite demand to LK. The previous method based on time-to-lane-crossing and cooperative states had its own drawbacks (could not re-plan a trajectory and did not actually provide supporting torque for LC) [22][24]. The Other solution, e.g., the wiper LC assistance which a driver could tell the ego-vehicle to change lane by switching a wiper could be potentially inconvenient and annoying [21], since it required additional operations and could not re-plan a new LC trajectory as well. In contrast, our method applied a deep learning-based method for detecting a driver's LC intention as a trigger to switch the system into the LC assistance mode [11]. Previous research showed similar applications in other human-machine interaction devices but employing intention detection in vehicle control is new [12][13]. Moreover, even though there is a possibility of launching the wrong assistance, the modified pseudo-work and gain turning method can allow the system to manage more situations without causing steering shaking, according to [24][31].

As for emergencies, e.g., avoiding leading vehicles or pedestrians, previous research has suggested the performance of the IBHS system. The IBHS system can only work in the following four situations in emergency avoidance:

(1) LC assistance inactivated but intention consistent
(2) LC assistance inactivated and intention inconsistent
(3) LC assistance activated and intention consistent
(4) LC assistance activated but intention inconsistent

In [32], the researchers explored the drivers' performance of avoiding leading car while driving with a haptic lane-keeping steering system. The drivers had to resist haptic torque to avoid a collision. The result suggested that, as long as the drivers held the steering wheel, the reaction time and the peak value of the steering angles remained almost consistent in comparison to the manual driving. Steering against a strong haptic torque slightly increased the collision risks in relative to manual driving and weak haptic assisting driving. The above scenario generally applies to the situation (1). It implies the Strong method IBHS system can play a role of resisting driver's steering and increase the possibility of collision. While a Weak method may not raise risk in comparison with manual driving. If steering without triggering LC assistance but with intention inconsistency, that applies to the situation (2). At first, the IBHS system may resist the driver to steer. But with the reduced torque strength, the steering authority can be smoothly transferred to the human driver. That will cause the driver to manually drive and avoid an obstacle.

In [39][40], the researchers tested the driving performance of evasive maneuvers while guided by haptic steering. Their conclusion illustrated that haptic steering could increase the mean minimum distance between the obstacle and improve safety. This applies to the situation (3). When the IBHS system captures a driver's intention, the LC assistance works as a guide to help the driver turns faster and increasing collision distance as [40]. If an inconsistency is detected in the avoiding process, that IBHS can guide the driver at the beginning then safely transfer into manual driving which applies to the situation (4).

The above discussion implies that the IBHS system is capable of handling emergency avoidance. Depending on the system intention detection result, the system can slightly increase the possibility of collisions or reduce the risks.

VI. CONCLUSION

An intention-based haptic guidance system that can switch between LK and LC assistance modes was realized via haptic guidance steering and a deep learning-based driver LC intention detection method. A driving simulator experiment compared driver performances under conditions of manual driving and assisted driving. The experiment explored two factors affecting assisting performance: the haptic strength and the LC speed. It can be seen from the driving simulator experiment that the Strong-Normal method was effective in shortening the LC duration, and showed a tendency to reduce the overshoot distance without causing steering shaking. Unlike previous research, a strong supporting torque showed better assistive performance than a weak torque when driving with IBHS assistance. The intention consistency detection method could accurately catch the driver's real intention and achieve smooth re-planning. In the future, more complicated traffic conditions will be explored, to further test the system.